\documentclass{article}

\usepackage{iclr2027_conference,times}
\usepackage{amsmath}
\usepackage{amssymb}
\usepackage{booktabs}
\usepackage{graphicx}
\usepackage{hyperref}
\usepackage{url}
\usepackage{wrapfig}
\usepackage{float}

\title{Self-Evolving Time-Series Forecasting Agents with Episodic Memory and \\Online Policy Learning}

\author{
Junyi Wang\textsuperscript{1,2} \quad
Yilin Wang\textsuperscript{2} \quad
Wen Wu\textsuperscript{2} \quad
Chao Zhang\textsuperscript{3,2} \\
\textsuperscript{1}Shanghai Jiao Tong University \\
\textsuperscript{2}Shanghai Artificial Intelligence Laboratory \\
\textsuperscript{3}Tsinghua University
}

\iclrfinalcopy

\begin{document}

\maketitle
\lhead{Preprint.}

\begin{abstract}

LLM-based agents are increasingly used for time-series forecasting because they can organise contextual information, perform multi-step analysis, and guide the sequence of actions required to complete forecasting tasks.
Most existing agents focus only on the current forecasting instance.
However, in real-world deployments, forecasting commonly operates online, with new forecasts issued from the currently available history as the forecast origin advances and the ground-truth targets of earlier instances progressively become available.
These targets provide feedback on the actions taken in earlier instances, yet existing agents generally do not preserve or utilise this information to adapt their subsequent actions.
To address this limitation, we introduce \textbf{FASE}, a \textbf{F}eedback-\textbf{A}ware \textbf{S}elf-\textbf{E}volving forecasting agent that converts such feedback into task-specific experience for subsequent forecasting instances.
FASE combines episodic memory, which retrieves relevant completed instances, with online policy learning, which summarises the feedback accumulated across instances into ranking guidance.
The proposed framework is evaluated on 29 dataset configurations selected from the GIFT-Eval benchmark.
Across these 29 configurations, FASE attains the strongest aggregate point forecasting performance among the evaluated methods and reduces the normalised MAE by 9.1\% relative to the best individual foundation model baseline.
The results further indicate that the cumulative advantage of FASE increases as delayed feedback accumulates.
Together, these findings demonstrate that FASE can continually self-evolve through feedback from completed forecasting instances without updating the parameters of the LLM.

\end{abstract}

\section{Introduction}

Time series forecasting supports a wide range of real-world applications, including power systems, transportation, and financial markets.
Recent advances in LLMs and agentic systems have inspired numerous studies that employ LLM-based agents for time-series forecasting.
By reasoning over contextual information and coordinating analytical and forecasting tools, these agents extend forecasting beyond a direct mapping from historical observations to future values \citep{tao2026castr1,ang2025tsagent}.
Existing work has focused only on the current forecasting instance.
However, in real-world deployments, forecasting is typically performed online over a sequence of instances.
Each forecast is based on the history available at that time, while the targets of earlier instances are revealed as time progresses.
These newly observed targets not only extend the history available for future forecasts but also reveal how well earlier actions of the agent performed.
Most forecasting agents use the delayed targets only as additional history, but rarely retain the feedback to improve subsequent actions.

To fill this gap, we introduce \textbf{FASE}, a \textbf{F}eedback-\textbf{A}ware \textbf{S}elf-\textbf{E}volving forecasting agent that makes feedback from completed forecasting instances available when later actions are selected.
FASE combines episodic memory with online policy learning to capture feedback at two levels.
Episodic memory incorporates each completed instance immediately and retrieves relevant instances as detailed evidence of tool performance under similar contexts.
Online policy learning periodically aggregates feedback across completed instances into a predicted ranking of forecasting tools for the current history.
For each forecasting instance, the LLM integrates the current time-series context with the retrieved evidence and ranking guidance to determine which forecasting tool to invoke.
When the corresponding ground-truth target becomes available, both components are updated so that the completed instance can influence later actions.

Across 29 dataset configurations selected from GIFT-Eval, FASE achieves the strongest aggregate point forecasting performance among all evaluated methods, including a 9.1\% reduction in normalised MAE relative to the best individual foundation model baseline.
The growth of these gains as delayed feedback accumulates provides empirical evidence that the agent self-evolves through experience from completed forecasting instances.

The main contributions of this work are summarised as follows.
\begin{itemize}
\item
We introduce \textbf{FASE}, a feedback-aware self-evolving forecasting agent that turns delayed feedback from completed forecasting instances into experience for later actions.
Without updating the LLM parameters, FASE enables continual adaptation of forecasting tool usage through accumulated experience.
\item
We develop a two-level feedback-driven adaptation mechanism.
Episodic memory retrieves relevant completed instances to provide context-specific evidence, while online policy learning periodically distils feedback accumulated across instances into ranking guidance for the current history.
Together, these components combine detailed evidence from individual instances with broader performance patterns learned from accumulated feedback.
\end{itemize}

\section{Related Work}

\subsection{Time-Series Foundation Models}

Time-series foundation models are pretrained on large and heterogeneous collections of time series to capture transferable temporal patterns, enabling zero-shot forecasting on previously unseen tasks.
Representative models include Chronos, TimesFM, Moirai, TimeGPT, Timer, MOMENT, and Time-MoE, which explore diverse architectures and pretraining strategies for scalable time-series modelling \citep{ansari2024chronos,das2023timesfm,woo2024moirai,garza2023timegpt,liu2024timer,goswami2024moment,shi2024timemoe}.
Recent studies have further extended this paradigm through continuous-valued generative pretraining in Sundial, billion-parameter serial scaling in Timer-S1, and sampling-rate-equivariant forecasting in FlowState \citep{liu2025sundial,liu2026timers1,graf2026flowstate}.
FASE builds upon these forecasting foundation models by treating them as numerical forecasting tools and investigating how delayed feedback from completed instances can improve future tool selection and usage.

\subsection{Time-Series Forecasting Agents}

Recent work has formulated forecasting as an agentic process in which an LLM reasons over contextual information and coordinates forecasting models and analytical tools.
KairosAgent combines an LLM-based reasoner with a time-series foundation model and dynamically invokes analytical tools to improve numerical understanding and semantic reasoning \citep{feng2026kairosagent}.
MemCast incorporates hierarchical memory derived from training data and adapts during inference by adjusting the confidence assigned to retrieved experiences \citep{tao2026memcast}.
EpiEvolve further explores feedback-driven adaptation by updating episodic experiences and strategic rules as delayed labels become available in a streaming forecasting scenario \citep{lu2026epievolve}.
However, its evaluation is conducted in a specific application domain with categorical forecasting targets.
Our work instead studies self-evolution through delayed feedback for numerical forecasting across heterogeneous time-series configurations.
FASE uses feedback from completed instances to continuously refine how the agent uses forecasting tools in future instances.

\begin{figure*}[t]
  \centering
  \includegraphics[width=\textwidth]{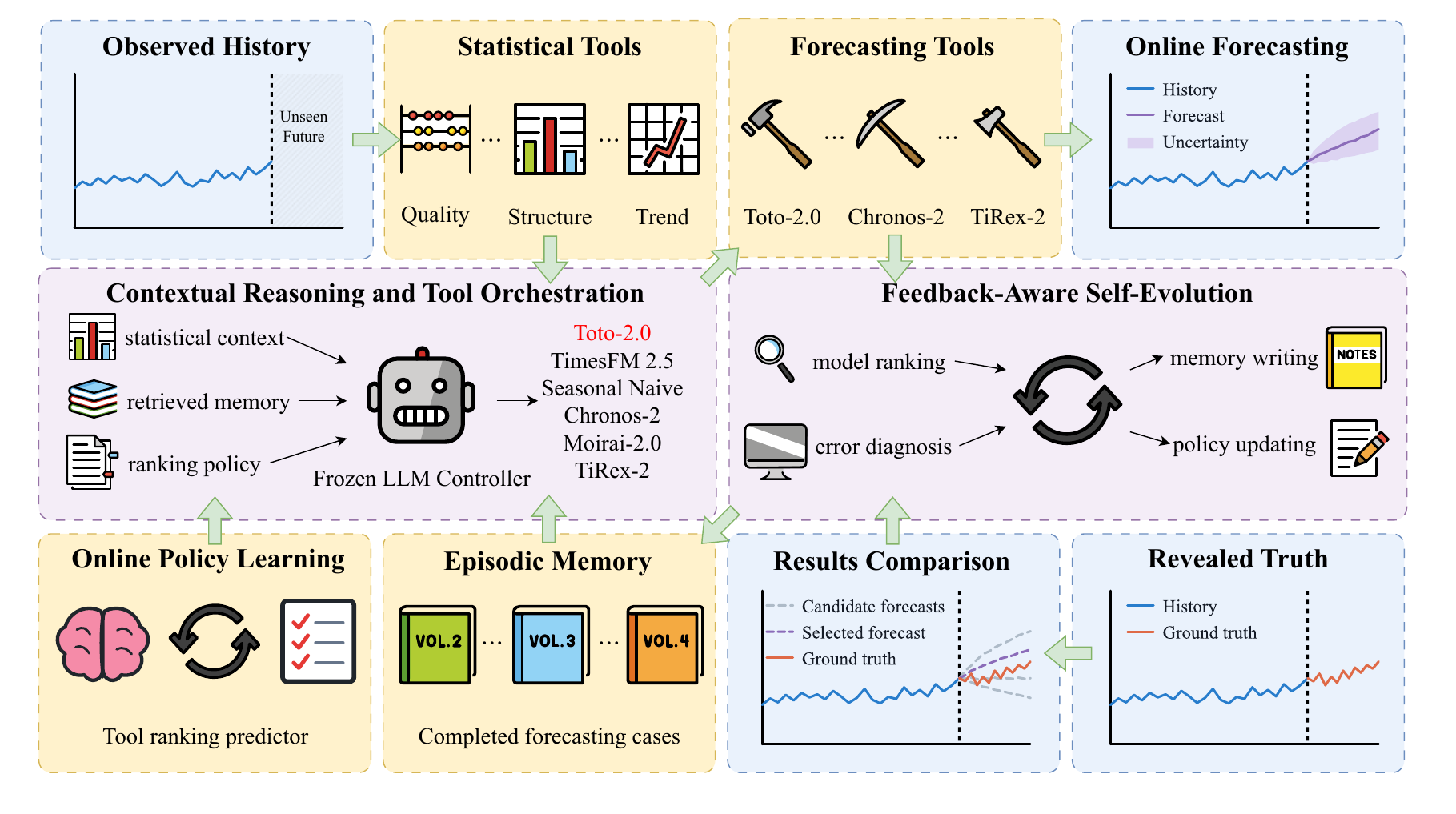}
  \caption{Overview of FASE. For each forecasting instance, the LLM uses statistical context, retrieved experience, and policy guidance to invoke a forecasting tool. After the ground-truth target is revealed, FASE compares the candidate forecasts with the target and uses the resulting feedback to update episodic memory and periodically update the online policy for subsequent actions.}
  \label{fig:overview}
\end{figure*}

\subsection{Self-Evolving LLM Agents}

Recent approaches enable LLM agents to improve over time by using feedback from prior interactions to update persistent external state.
Reflexion converts environmental feedback into verbal reflections, while ExpeL extracts reusable knowledge from accumulated experience \citep{shinn2023reflexion,zhao2024expel}.
More recent methods extend this idea through other forms of adaptation.
ReasoningBank distils generalisable strategies from successful and failed experiences, ACE incrementally updates a contextual playbook, and GEPA uses reflection to optimise prompt components \citep{ouyang2026reasoningbank,zhang2026ace,agrawal2026gepa}.
These approaches primarily focus on adapting textual knowledge, reusable strategies, or prompt-level behaviours.
FASE extends this paradigm to delayed numerical feedback in online forecasting.
Specifically, episodic memory preserves completed forecasting instances for retrieval according to the current context, whereas online policy learning periodically consolidates feedback across instances into a predicted ranking of forecasting tools.

\section{FASE}
\label{sec:method}

\subsection{Overview}

FASE operates over a sequence of forecasting instances with access to a fixed library of forecasting tools.
To use information from completed instances in subsequent actions, FASE maintains two forms of external state: episodic memory and an online policy.
Both are initialised empty for each task configuration and updated as ground-truth targets become available.
Figure~\ref{fig:overview} summarises how FASE processes each forecasting instance and uses delayed feedback to update episodic memory and the online policy.

For each forecasting instance, the forecasting tools and the LLM receive different representations of the observed history.
Each forecasting tool operates directly on the original numerical values.
In contrast, providing raw numerical values to the LLM requires the time series to be serialised into a token sequence, resulting in substantial input overhead for long histories containing thousands of observations.
Moreover, serialised numerical values provide limited explicit semantic information, making it difficult for the LLM to directly interpret the temporal patterns and structural characteristics of the time series.
FASE therefore represents the observed history using a fixed collection of statistical features, which summarise properties such as scale, trend, periodicity, temporal dependence, and missingness.
This representation provides the LLM with a compact and structured description of the time series while avoiding the token overhead associated with representing the full numerical history.

Based on the current statistical representation, episodic memory retrieves relevant completed forecasting instances.
After the warm-up phase, the online policy additionally produces a ranking of candidate tools.
FASE provides the LLM with the task context, statistical representation, retrieved experience, available policy guidance, and descriptions of the forecasting tools.
The LLM selects a single tool to produce the point forecast and corresponding quantile forecasts for the current forecasting instance.
The remaining tools are run separately on the same observed history before the target is revealed, and all forecasts are stored for evaluation when the target becomes available.
After the ground-truth target is revealed, FASE computes the point and probabilistic forecast errors of all tools.
The statistical representation, task context, invoked tool, and observed errors form the record of a completed forecasting instance.
This record enters episodic memory and supplies feedback to the online policy.

\subsection{Episodic Memory}

After the ground-truth target for a forecasting instance becomes available, FASE organises the statistical representation, task context, invoked tool, and observed errors into an episodic memory entry for subsequent retrieval.
For each candidate tool, the entry records the point forecast error and its corresponding rank, together with the probabilistic forecast error as auxiliary feedback.

FASE organises the active memory, which contains all entries currently eligible for retrieval, into a recent pool and a long-term pool.
After a forecasting instance is completed, FASE first updates the retention values of the memory entries included in the context used to select the corresponding action and then inserts the newly completed instance directly into the recent pool.
If this insertion causes the recent pool to exceed its capacity, FASE removes the oldest entry and considers it for promotion to the long-term pool.
If the long-term pool has available capacity, the entry is promoted directly.
Otherwise, whether the candidate enters the long-term pool is determined by its retention value.

Let $s_i$ denote the statistical representation of forecasting instance $i$.
To compute retention values, FASE measures the similarity between the current representation $s_i$ and the representation $s_j$ stored in an active entry using a normalised statistical distance.
Let $\mathcal{G}_{ij}$ denote the statistical feature groups containing observed features in both $s_i$ and $s_j$, and let $\mathcal{Q}_{ij,g}$ denote the corresponding features in group $g$.
The distance is defined as
\begin{equation}
d(s_i,s_j)=\frac{1}{|\mathcal{G}_{ij}|}
\sum_{g\in\mathcal{G}_{ij}}
\frac{1}{|\mathcal{Q}_{ij,g}|}
\sum_{q\in\mathcal{Q}_{ij,g}}
\rho\!\left(\frac{|s_{i,q}-s_{j,q}|}{\sigma_q}\right),
\qquad
\rho(z)=\frac{z}{1+z},
\label{eq:memory-distance}
\end{equation}
where $\sigma_q$ is the standard deviation of feature $q$ across active entries.
The transformation $\rho$ bounds the influence of features with large normalised discrepancies.

For each forecasting instance, active entries are partitioned according to the tool used in each stored instance and the set of candidate tools tied for the best rank based on point forecast error.
Let $\mathcal{E}_i$ denote the entries included in the context used to select the action for instance $i$.
For each partition $C$, let $j_C^{(1)}$ and $j_C^{(2)}$ denote the nearest and second-nearest entries, respectively.
The contribution of entry $j$ is defined as
\begin{equation}
\delta_{i,j}=
\begin{cases}
d(s_i,s_{j_C^{(2)}})-d(s_i,s_{j_C^{(1)}}), & j=j_C^{(1)}\in\mathcal{E}_i,\ |C|>1, \\
1-d(s_i,s_{j_C^{(1)}}), & j=j_C^{(1)}\in\mathcal{E}_i,\ |C|=1, \\
0, & \text{otherwise}.
\end{cases}
\label{eq:retention-contribution}
\end{equation}
where $C$ denotes the partition containing entry $j$.
The retention value of each active entry is the running average of its contributions across retention updates.
Thus, a large retention value indicates that the entry is consistently a more distinctive representative of the associated tool use and observed candidate performance than nearby alternatives.

Given the current statistical representation $s_i$, FASE ranks all active entries according to $d(s_i,s_j)$, retrieves those with the smallest distances, and presents their contextual statistics and observed tool performance to the LLM as evidence for selecting the current action.

\subsection{Online Policy Learning}

Episodic memory incorporates feedback from each completed forecasting instance immediately and makes relevant instances available when subsequent actions are selected.
Online policy learning instead aggregates feedback over multiple completed instances and periodically updates the policy state based on accumulated experience.
FASE implements this policy state using a lightweight predictor $g_{\theta_i}$, which is updated online, and a first-in, first-out feedback pool.

Let $\mathcal{F}$ denote the set of available forecasting tools.
Let $x_i\in\mathbb{R}^{N}$ denote the $N$ most recent visible values in the current history.
FASE normalises the observed values and combines them with a binary observation mask, producing $z_i=[\operatorname{Normalise}(x_i),\operatorname{Mask}(x_i)]\in\mathbb{R}^{2N}$.
The predictor is a lightweight MLP $g_{\theta_i}:\mathbb{R}^{2N}\rightarrow\mathbb{R}^{|\mathcal{F}|}$ that maps $z_i$ to relative performance scores $u_i=g_{\theta_i}(z_i)$.
A lower score $u_{i,m}$ indicates better expected point forecasting performance for tool $m$.
Sorting the scores in ascending order produces a tool ranking, which FASE converts into a textual list and provides to the LLM as additional policy guidance for the current action.

When the ground-truth target for forecasting instance $i$ becomes available, FASE adds the input representation and observed tool errors to the feedback pool.
After sufficient feedback has accumulated, the predictor is initialised using the available feedback pool and subsequently updated periodically as additional instances are completed.
Each update samples minibatches uniformly from the current feedback pool.
The corresponding hyperparameters are given in Section~\ref{sec:implementation-details}.
For a record $(z,\ell)$ drawn from such a minibatch, let $u=g_\theta(z)$ and $\bar{\ell}=|\mathcal{F}|^{-1}\sum_{m\in\mathcal{F}}\ell_m$.
The predictor minimises the following pairwise loss, averaged over all tool pairs and then over the minibatch:
\begin{equation}
\mathcal{L}(z,\ell)=
\frac{1}{\binom{|\mathcal{F}|}{2}}
\sum_{p<q}
\frac{|\ell_p-\ell_q|}{\bar{\ell}}
\operatorname{softplus}\!\left(
\operatorname{sgn}(\ell_q-\ell_p)(u_p-u_q)
\right),
\label{eq:ranking-loss}
\end{equation}
The sign term orients each pair so that the tool with lower point error is encouraged to receive the lower score.
The loss assigns larger weights to pairs with greater relative differences in point forecast error.

\section{Experiments}
\label{sec:experiments}

\subsection{Datasets and Metrics}

We evaluate FASE on 29 dataset configurations selected from the 97 configurations in GIFT-Eval \citep{aksu2024gift}, retaining those with 500 to 5,000 forecasting instances in the test split.
The lower bound provides enough forecasting instances to examine self-evolution as feedback accumulates, while the upper bound limits computational cost.
The selected configurations cover all seven GIFT-Eval domains: Econ/Fin, Energy, Healthcare, Nature, Sales, Transport, and Web/CloudOps.
Some source datasets contribute multiple configurations under different sampling frequencies or forecast terms.
Together, the configurations cover sampling frequencies from 10 seconds to one month, short, medium, and long forecast terms, and prediction lengths ranging from 8 to 720 time steps.
Detailed information about the evaluated datasets is provided in Appendix~\ref{app:dataset-configurations}.

We follow the GIFT-Eval protocol for test splits and the construction of forecasting instances.
Each configuration may include multiple series, with up to 20 forecasting instances per series.
Episodic memory and the online policy accumulate at the task level rather than per series: the agent processes all forecasting instances of one series in chronological order, then moves to the next series, following the item order used by GIFT-Eval.

We evaluate point forecasts using MASE, MAE, and RMSE, and probabilistic forecasts using CRPS.
CRPS is estimated from the nine predicted quantiles at levels 0.1 through 0.9.
For every task configuration, each metric is normalised by the corresponding score of Seasonal Naive.
We report the geometric mean of these normalised scores across the 29 configurations, giving each configuration equal weight.
For the rank columns in Table~\ref{tab:main-results}, systems are ranked within each configuration for each metric.
Tied systems receive the mean of their ranks, and the resulting ranks are then averaged arithmetically across configurations.

\subsection{Implementation Details}
\label{sec:implementation-details}

FASE uses a fixed tool pool that comprises Seasonal Naive and five time-series foundation models: Chronos-2~\citep{ansari2025chronos2}, TimesFM~2.5~\citep{google2025timesfm25}, Moirai-2.0-R-Small~\citep{liu2025moirai2}, TiRex-2~\citep{podest2026tirex2}, and Toto-2.0-313m~\citep{khwaja2026toto2}.
These forecasters also serve as the individual model baselines.
To isolate the effects of episodic memory and online policy learning, we define Base Agent as an additional baseline.
Its LLM receives only the context of the current forecasting instance, including the statistical summary and descriptions of the available forecasting tools, without retrieved instances or policy guidance.
Qwen3.8-27B-FP8~\citep{qwen2026qwen38} serves as the LLM controller, with the temperature set to 0.1, top-p set to 1.0, and the reasoning effort set to xhigh to support consistent and deliberate action selection.
The prompt templates used by the controller are provided in Appendix~\ref{app:prompt-templates}.

For each forecasting instance, the observed history is summarised by the 18 statistical features that define $s_i$.
This representation is provided to the LLM as a compact description of the current history and is also stored with each episodic memory entry.
The features characterise temporal dynamics, missingness, periodicity, covariate relationships, and other properties of the history.
Appendix~\ref{app:statistical-features} lists the individual definitions of these features.

The recent and long-term memory pools have capacities of 100 and 900 entries, respectively, and FASE retrieves ten entries for each forecasting instance.
The online policy instead operates on the most recent $8H$ history values, which are normalised separately for each forecasting instance and paired with a binary observation mask.
Completed instances are inserted into a first-in, first-out feedback pool with a capacity of 3,000 records.
The ranker is initialised after feedback from 100 instances has been collected, using 10 epochs, and is subsequently updated after every 25 additional valid instances, using 25 minibatches of 256 uniformly sampled records.
For optimisation, we use AdamW \citep{loshchilov2019adamw} with a learning rate of $10^{-3}$ and a weight decay of $10^{-5}$.

\section{Results and Discussion}
\label{sec:results}

\subsection{Main Results}

Table~\ref{tab:main-results} compares FASE with the fixed forecasting tools and with Uniform Ensemble over the 29 evaluated GIFT-Eval configurations.
Appendix~\ref{app:per-configuration-results} reports the corresponding results for each configuration.
FASE achieves the strongest overall point forecasting performance, with the lowest normalised errors for MASE, MAE, and RMSE.
TiRex-2 is the strongest individual foundation model on all three point metrics.
Relative to TiRex-2, FASE reduces MASE, MAE, and RMSE by 6.8\%, 9.1\%, and 5.0\%, respectively.
The improvements across complementary error measures show that the advantage of FASE is not tied to a single definition of point forecast quality.

\begin{table*}[t]
\centering
\small
\setlength{\tabcolsep}{3.2pt}
\caption{Aggregate results over the 29 evaluated GIFT-Eval configurations. Bold and underlined values indicate the best and second-best results within each column, respectively.}
\label{tab:main-results}
\begin{tabular*}{\textwidth}{@{\extracolsep{\fill}}lcccccccc@{}}
\toprule
 & \multicolumn{2}{c}{MASE $\downarrow$} & \multicolumn{2}{c}{MAE $\downarrow$} & \multicolumn{2}{c}{RMSE $\downarrow$} & \multicolumn{2}{c}{CRPS $\downarrow$} \\
\cmidrule(lr){2-3} \cmidrule(lr){4-5} \cmidrule(lr){6-7} \cmidrule(lr){8-9}
Method & Norm. & Rank & Norm. & Rank & Norm. & Rank & Norm. & Rank \\
\midrule
Seasonal Naive & 1.000 & 7.79 & 1.000 & 7.72 & 1.000 & 7.86 & 1.000 & 7.97 \\
Chronos-2~\citep{ansari2025chronos2} & 0.752 & 4.55 & 0.729 & 5.03 & 0.720 & 4.59 & 0.545 & 4.90 \\
TimesFM~2.5~\citep{google2025timesfm25} & 0.753 & 4.59 & 0.737 & 4.62 & 0.705 & 3.86 & 0.528 & 4.17 \\
Moirai-2.0-R-Small~\citep{liu2025moirai2} & 0.774 & 5.97 & 0.739 & 5.21 & 0.756 & 6.24 & 0.542 & 5.38 \\
TiRex-2~\citep{podest2026tirex2} & \underline{0.728} & 4.34 & \underline{0.685} & 3.97 & \underline{0.677} & 3.72 & \textbf{0.489} & 3.38 \\
Toto-2.0-313m~\citep{khwaja2026toto2} & 0.745 & 4.03 & 0.699 & 4.34 & 0.716 & 4.93 & 0.500 & 3.93 \\
Uniform Ensemble\textsuperscript{a} & 0.731 & \underline{2.86} & 0.697 & \underline{3.41} & 0.681 & \underline{2.86} & 0.501 & \textbf{3.10} \\
\midrule
FASE & \textbf{0.679} & \textbf{1.86} & \textbf{0.622} & \textbf{1.69} & \textbf{0.643} & \textbf{1.93} & \underline{0.497} & \underline{3.17} \\
\bottomrule
\end{tabular*}
\par\vspace{2pt}
\begin{minipage}{\textwidth}
\footnotesize\raggedright
\textsuperscript{a} Uniform Ensemble averages the forecasts of the five foundation models with equal weights.
\end{minipage}
\end{table*}

The average ranks provide a complementary view by comparing systems within each configuration and then weighting all configurations equally.
FASE also ranks first on all three point metrics, showing that its aggregate gains are accompanied by consistently strong relative performance rather than being driven by a small number of configurations with unusually large improvements.
FASE also outperforms Uniform Ensemble, the strongest fixed baseline by average rank, reducing its normalised MASE, MAE, and RMSE by 7.1\%, 10.8\%, and 5.6\%, respectively.
This improvement over a fixed equal-weight combination indicates that FASE learns from accumulated feedback which forecasting tool is better suited to each forecasting instance.

The gains are less consistent for probabilistic forecasting.
FASE obtains the second-lowest aggregate CRPS of 0.497, behind TiRex-2 at 0.489, while its average CRPS rank is also slightly higher than that of Uniform Ensemble.
This difference is consistent with the feedback mechanism that guides tool selection.
The retrieved instances emphasise point forecasting errors and the associated ranks, and the online policy is optimised to order tools according to point forecasting performance.
Probabilistic error is incorporated only as auxiliary feedback.
Accordingly, FASE delivers more consistent gains in point forecasting than in probabilistic forecasting.

Table~\ref{tab:horizon-results} examines whether the point forecasting advantage is concentrated at a particular prediction length.
FASE obtains the lowest normalised MASE in the long, medium, and short groups, reducing MASE relative to TiRex-2 by approximately 6.0\%, 6.2\%, and 7.3\%, respectively.
The similar gains across the three groups show that the aggregate improvement persists over the evaluated range of prediction lengths.
The CRPS results are less consistent across groups: FASE performs best on the short configurations and second best on the medium configurations, but does not lead on the long configurations.
This contrast again shows that the main benefit of the current feedback design lies in point forecasting.

\begin{table*}[t]
\centering
\small
\setlength{\tabcolsep}{3pt}
\caption{Results aggregated by prediction length. Bold and underlined values indicate the best and second-best results within each column, respectively.}
\label{tab:horizon-results}
\begin{tabular*}{\textwidth}{@{\extracolsep{\fill}}lcccccc@{}}
\toprule
 & \multicolumn{2}{c}{Long} & \multicolumn{2}{c}{Medium} & \multicolumn{2}{c}{Short} \\
\cmidrule(lr){2-3} \cmidrule(lr){4-5} \cmidrule(lr){6-7}
Method & MASE $\downarrow$ & CRPS $\downarrow$ & MASE $\downarrow$ & CRPS $\downarrow$ & MASE $\downarrow$ & CRPS $\downarrow$ \\
\midrule
Seasonal Naive & 1.000 & 1.000 & 1.000 & 1.000 & 1.000 & 1.000 \\
Chronos-2~\citep{ansari2025chronos2} & 0.782 & 0.547 & 0.782 & 0.563 & 0.729 & 0.538 \\
TimesFM~2.5~\citep{google2025timesfm25} & 0.763 & 0.519 & 0.773 & 0.546 & 0.741 & 0.525 \\
Moirai-2.0-R-Small~\citep{liu2025moirai2} & 0.823 & 0.560 & 0.831 & 0.567 & 0.733 & 0.525 \\
TiRex-2~\citep{podest2026tirex2} & \underline{0.748} & \underline{0.475} & \underline{0.760} & \textbf{0.488} & \underline{0.708} & \underline{0.496} \\
Toto-2.0-313m~\citep{khwaja2026toto2} & 0.753 & \textbf{0.474} & 0.786 & 0.500 & 0.727 & 0.512 \\
Uniform Ensemble & 0.749 & 0.490 & 0.762 & 0.509 & 0.711 & 0.503 \\
\midrule
FASE & \textbf{0.703} & 0.515 & \textbf{0.713} & \underline{0.492} & \textbf{0.656} & \textbf{0.491} \\
\bottomrule
\end{tabular*}
\end{table*}

\subsection{Self-Evolution over Accumulated Feedback}

\begin{figure}[t]
  \centering
  \includegraphics[width=0.60\linewidth]{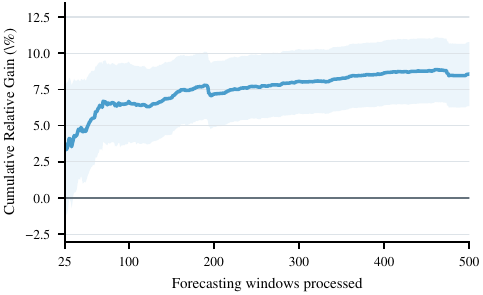}
  \caption{Cumulative relative MASE gain of FASE over the Base Agent across the first 500 forecasting instances in the replay. The curve begins at 25 processed instances, and the shaded band denotes the standard error across configurations.}
  \label{fig:self-evolution}
\end{figure}

We examine whether FASE can continually self-evolve through episodic memory and online policy learning.
For each configuration, we calculate the mean MASE of FASE and the Base Agent over all forecasting instances processed up to each point.
The reduction achieved by FASE is expressed as a percentage of the Base Agent MASE and then averaged equally across configurations.
Figure~\ref{fig:self-evolution} presents the resulting trajectory over the first 500 processed instances.

The cumulative gain exhibits an overall upward trend, rising from 3.8\% after 25 processed instances to 6.7\% after 100 instances and 8.6\% after 500 instances.
Despite minor local fluctuations, the relative advantage remains positive and continues to increase as more forecasting instances are completed.
Because each point includes all preceding forecasting instances, this sustained rise shows that later instances strengthen rather than dilute the advantage accumulated earlier.
The trend provides evidence that the feedback incorporated into episodic memory and the online policy continues to improve subsequent actions as experience grows.

\subsection{Ablation Study}

Table~\ref{tab:ablation} evaluates the separate and joint contributions of episodic memory and online policy learning.
Memory Only and Policy Only both improve every reported metric over the Base Agent.
In particular, they reduce normalised MASE from 0.758 to 0.683 and 0.687, respectively, showing that each mechanism provides useful guidance even when used independently.
Episodic memory contributes evidence from completed instances with contexts similar to the current one, whereas the online policy periodically consolidates feedback from a broader set of instances into a ranking of the forecasting tools.
Their similar point forecasting performance suggests that these two forms of feedback are individually useful despite operating at different levels of aggregation.

FASE combines both mechanisms and achieves the lowest MASE, MAE, and RMSE among the four variants, with relative reductions ranging from 9.8\% to 13.4\% relative to the Base Agent.
It also performs consistently better on all three point metrics than either variant using only one mechanism.
Although the additional gains are modest, their consistency supports the complementary roles of evidence retrieved from similar contexts and broader performance patterns captured by the policy.
Memory Only retains the lowest CRPS, while FASE ranks second on this metric.
This exception is consistent with the design of both feedback mechanisms, which prioritise point forecasting errors and use probabilistic error only as auxiliary information.

\begin{table*}[t]
\centering
\small
\setlength{\tabcolsep}{3.2pt}
\caption{Ablation of episodic memory and online policy learning over the same 29 evaluated GIFT-Eval configurations. Base Agent uses neither mechanism; Memory Only uses episodic memory; Policy Only uses online policy learning; FASE combines both. Bold and underlined values indicate the best and second-best results within each column, respectively.}
\label{tab:ablation}
\begin{tabular*}{0.7\textwidth}{@{\extracolsep{\fill}}lcccc@{}}
\toprule
Method & MASE $\downarrow$ & MAE $\downarrow$ & RMSE $\downarrow$ & CRPS $\downarrow$ \\
\midrule
Base Agent & 0.758 & 0.718 & 0.713 & 0.543 \\
Memory Only & \underline{0.683} & 0.631 & 0.655 & \textbf{0.483} \\
Policy Only & 0.687 & \underline{0.629} & \underline{0.649} & 0.505 \\
\midrule
FASE & \textbf{0.679} & \textbf{0.622} & \textbf{0.643} & \underline{0.497} \\
\bottomrule
\end{tabular*}
\end{table*}

\begin{figure}[t]
  \centering
  \includegraphics[width=0.75\textwidth]{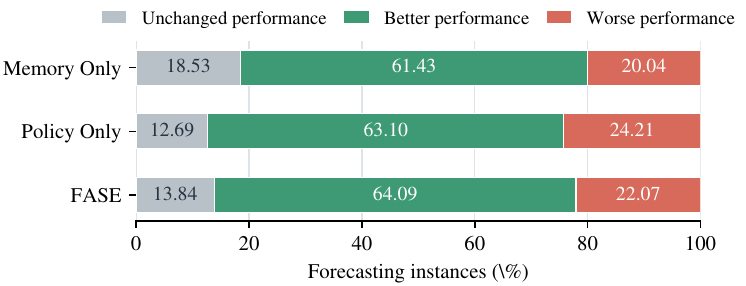}
  \caption{Distribution of forecasting performance relative to the Base Agent. Better and worse performance correspond to lower and higher MASE, respectively. Unchanged performance includes forecasting instances in which the evaluated method selects the same tool as the Base Agent or selects a different tool that yields the same MASE; the latter account for less than 0.3\% of all evaluated instances. Results are averaged equally across configurations.}
  \label{fig:memory-policy-analysis}
\end{figure}

\begin{table}[t]
  \centering
  \small
  \setlength{\tabcolsep}{3.2pt}
  \renewcommand{\arraystretch}{1.18}
  \caption{MASE changes relative to the Base Agent. Improvement is the MASE reduction from forecasting instances with better performance. Degradation is defined analogously for instances with worse performance. Net improvement is their difference. Results are then averaged equally across configurations.}
  \label{tab:memory-policy-analysis}
  \begin{tabular*}{0.65\textwidth}{@{\extracolsep{\fill}}lccc@{}}
    \toprule
    & \multicolumn{3}{c}{MASE change relative to Base Agent (\%)} \\
    \cmidrule(lr){2-4}
    Method & Improvement & Degradation & Net improvement \\
    \midrule
    Memory Only & 11.4 & 2.2 & 9.2 \\
    Policy Only & 11.7 & 2.6 & 9.1 \\
    FASE & 11.9 & 2.2 & 9.7 \\
    \bottomrule
  \end{tabular*}
\end{table}

\subsection{Analysis of Memory and Policy}

Episodic memory influences tool use by retrieving completed instances with contexts similar to the current forecasting instance and presenting the observed performance of each forecasting tool in those instances.
The LLM combines this evidence with the current context to assess whether a previous tool advantage is likely to persist.
As shown in Figure~\ref{fig:memory-policy-analysis}, Memory Only performs better than the Base Agent on 61.4\% of forecasting instances and worse on 20.0\%, indicating that local performance patterns often transfer to similar contexts but do not always persist.
Online policy learning provides a more direct signal: the ranker converts accumulated feedback into a predicted ordering of the tools, which the LLM considers when choosing its action.
The policy therefore improves forecasting performance when this ordering captures the current dynamics, but can degrade performance when the learned pattern does not apply.

FASE jointly uses these two sources of information.
Table~\ref{tab:memory-policy-analysis} shows that the resulting improvement of 11.9\% outweighs the degradation of 2.2\%, yielding the largest net MASE improvement of 9.7\%.
Compared with the variants using only one mechanism, FASE achieves a larger aggregate improvement without increasing degradation beyond the level observed for Memory Only.
The positive balance shows how combining evidence from similar completed instances with the broader ordering learned by the policy improves tool use overall, even though either signal can occasionally lead to a worse action.

\section{Conclusion}

LLM-based agents are increasingly used for time-series forecasting, with existing approaches focusing only on the current forecasting instance.
In real-world deployments, however, forecasting commonly operates online, and the ground-truth targets of earlier instances become available over time.
Existing agents rarely retain the feedback provided by these targets to improve subsequent actions.
To address this gap, we introduced FASE, a feedback-aware self-evolving forecasting agent that combines episodic memory with online policy learning.
Episodic memory retrieves relevant experience from completed forecasting instances, while online policy learning summarises accumulated feedback into ranking guidance for subsequent actions.
Experiments on 29 GIFT-Eval configurations show that FASE achieves the strongest aggregate point forecasting performance among the evaluated methods.
Its performance gains increase as feedback accumulates, and further analysis shows that memory and policy guidance change how the agent uses forecasting tools in subsequent instances.
Together, these findings demonstrate that FASE can continually self-evolve through feedback from completed forecasting instances.

\section*{AI use statement}

In this work, we used generative AI tools for drafting a section of this manuscript, assisting with writing and language polishing, identifying related work, analysing experimental data, and interpreting the reported results.
We have reviewed all AI-assisted work. We take responsibility for the final content of this work, including text, claims, or artefacts produced with the aid of generative AI.

\bibliographystyle{iclr2027_conference}
\bibliography{references}

\clearpage
\appendix

\section{Evaluation Configurations}
\label{app:dataset-configurations}

Table~\ref{tab:dataset-configurations} provides detailed information about the 29 GIFT-Eval configurations used in our evaluation.
Configurations from the same source dataset are grouped together.

\begin{table}[H]
\centering
\scriptsize
\caption{GIFT-Eval configurations used for evaluation. Rows are grouped by source dataset. Prediction length is measured in time steps.}
\label{tab:dataset-configurations}
\begin{tabular}{llclrr}
\toprule
Dataset & Domain & Frequency & Term & Pred. Length & Instances \\
\midrule
bitbrains\_fast\_storage & Web/CloudOps & 5T & medium & 480 & 5,000 \\
bitbrains\_fast\_storage & Web/CloudOps & 5T & long & 720 & 5,000 \\
bitbrains\_fast\_storage & Web/CloudOps & H & short & 48 & 5,000 \\
\addlinespace
bitbrains\_rnd & Web/CloudOps & 5T & medium & 480 & 2,000 \\
bitbrains\_rnd & Web/CloudOps & 5T & long & 720 & 2,000 \\
bitbrains\_rnd & Web/CloudOps & H & short & 48 & 2,000 \\
\addlinespace
bizitobs\_service & Web/CloudOps & 10S & short & 60 & 630 \\
\addlinespace
car\_parts\_with\_missing & Sales & M & short & 12 & 2,674 \\
\addlinespace
electricity & Energy & D & short & 30 & 1,850 \\
electricity & Energy & H & medium & 480 & 2,960 \\
electricity & Energy & H & long & 720 & 1,850 \\
electricity & Energy & W-FRI & short & 8 & 1,110 \\
\addlinespace
hierarchical\_sales & Sales & D & short & 30 & 826 \\
\addlinespace
hospital & Healthcare & M & short & 12 & 767 \\
\addlinespace
kdd\_cup\_2018\_with\_missing & Nature & D & short & 30 & 540 \\
kdd\_cup\_2018\_with\_missing & Nature & H & medium & 480 & 540 \\
kdd\_cup\_2018\_with\_missing & Nature & H & long & 720 & 540 \\
\addlinespace
LOOP\_SEATTLE & Transport & 5T & long & 720 & 4,845 \\
LOOP\_SEATTLE & Transport & D & short & 30 & 646 \\
LOOP\_SEATTLE & Transport & H & medium & 480 & 646 \\
LOOP\_SEATTLE & Transport & H & long & 720 & 646 \\
\addlinespace
m4\_daily & Econ/Fin & D & short & 14 & 4,227 \\
\addlinespace
M\_DENSE & Transport & H & short & 48 & 600 \\
\addlinespace
restaurant & Sales & D & short & 30 & 807 \\
\addlinespace
solar & Energy & 10T & short & 48 & 2,740 \\
solar & Energy & 10T & medium & 480 & 1,507 \\
solar & Energy & 10T & long & 720 & 1,096 \\
solar & Energy & H & short & 48 & 2,603 \\
\addlinespace
SZ\_TAXI & Transport & 15T & short & 48 & 1,092 \\
\bottomrule
\end{tabular}
\end{table}

\clearpage
\section{Online Forecasting Protocol}
\label{app:online-protocol}

We evaluate all methods by replaying the forecasting instances of each GIFT-Eval configuration in a fixed order.
Within each series, the instances are processed chronologically; after all instances of one series have been processed, evaluation proceeds to the next series in the item order provided by GIFT-Eval.
Because the benchmark does not define a shared clock across different series, no additional temporal ordering is imposed between series.
Episodic memory and the online policy persist across series within the same configuration, but both are reinitialised for every new configuration.

At each forecast origin, tool inputs are constructed only from the information available at that time.
All evaluated forecasting instances are univariate.
GIFT-Eval configurations with multivariate targets are converted into separate univariate series before inference, so every forecasting tool receives a single target channel.
Seasonal Naive uses the complete available history and a seasonal period inferred from the sampling frequency.
For Chronos-2, TimesFM~2.5, Moirai-2.0-R-Small, TiRex-2, and Toto-2.0-313m, we retain the most recent 8,192, 15,360, 4,000, 8,192, and 4,096 time steps, respectively.
Where available, past and known future covariates are supplied to TiRex-2, while Toto-2.0-313m receives past covariates; the remaining tools receive only the target history.
Chronos-2, Moirai-2.0-R-Small, TiRex-2, and Toto-2.0-313m receive missing target values directly through their native interfaces.
For Seasonal Naive and TimesFM~2.5, missing values are linearly interpolated over timestamps within each target channel.
Leading and trailing missing values are replaced by the nearest observed value, and a channel with no observed values is filled with zeros.

The LLM chooses its action using the current statistical context and any memory or policy information derived from previously completed instances.
The selected tool produces the forecast used by the agent, while the remaining tools are run separately on the same available history before the target is revealed.
The candidate forecasts are not provided to the LLM, and none of the tools has access to the target during inference.
TimesFM~2.5 is run with input normalisation, positivity inference, flip invariance, a continuous quantile head, and quantile crossing correction enabled.
Each adapter returns a point forecast and nine quantile forecasts at levels $0.1$ through $0.9$, all arranged in the same horizon-by-channel format.
No additional scaling, clipping, or calibration is applied after the adapter output.
Uniform Ensemble is constructed by averaging the point forecasts of the five foundation models with equal weights and applying the same averaging operation separately at each quantile level.

After the target becomes available, we compute the errors of all six candidate forecasts.
The resulting feedback is then added to episodic memory and to the feedback pool of the online policy, and any state update takes effect from the next forecasting instance.

\clearpage
\section{Complete Implementation Settings}
\label{app:implementation-settings}

Table~\ref{tab:implementation-settings} summarises the implementation settings used in all reported experiments.
The online policy is maintained independently for each GIFT-Eval configuration, as is episodic memory.

\begin{table}[H]
\centering
\small
\setlength{\tabcolsep}{5pt}
\caption{Complete implementation settings used in the experiments.}
\label{tab:implementation-settings}
\begin{tabular}{p{0.25\textwidth}p{0.65\textwidth}}
\toprule
Component & Setting \\
\midrule
LLM controller & Qwen3.8-27B-FP8; temperature $0.1$; top-p $1.0$; reasoning effort xhigh; sampling seed $1234$. \\
Statistical context & 18 features computed from up to the 15,360 most recent history positions. \\
Episodic memory & Recent pool capacity $100$; long-term pool capacity $900$; $10$ retrieved instances per action. \\
Policy input & Most recent $8H$ observations, normalised within each forecasting instance using the median and standard deviation, together with a binary observation mask. \\
Policy predictor & MLP with dimensions $2(8H)\rightarrow 2048\rightarrow 128\rightarrow 6$ and ReLU activations after the hidden layers. \\
Feedback pool & First-in, first-out pool with capacity $3{,}000$; uniform minibatch sampling. \\
Initial policy training & Begins after $100$ completed instances and runs for $10$ epochs. \\
Periodic policy update & Performed after every $25$ additional completed instances using $25$ minibatches of size $256$. \\
Policy optimisation & AdamW; learning rate $10^{-3}$; weight decay $10^{-5}$; random seed $7$. \\
\bottomrule
\end{tabular}
\end{table}

\clearpage
\section{Statistical Features}
\label{app:statistical-features}

Table~\ref{tab:statistical-features} defines the 18 features used to summarise the observed history of each forecasting instance.

\begin{table}[H]
\centering
\scriptsize
\caption{Statistical features used to form the representation $s_i$.}
\label{tab:statistical-features}
\begin{tabular}{p{0.16\textwidth}p{0.18\textwidth}p{0.50\textwidth}}
\toprule
Group & Statistic & Description \\
\midrule
Data quality & Missing ratio & Fraction of missing target values. \\
Data quality & Zero ratio & Fraction of observed target values equal to zero. \\
\addlinespace
Temporal dynamics & Global trend strength & Standardised slope over the full history. \\
Temporal dynamics & Recent trend strength & Standardised slope over the recent history. \\
Temporal dynamics & Recent level shift strength & Standardised shift between recent and prior means. \\
Temporal dynamics & Recent scale change strength & Normalised change between recent and prior scales. \\
\addlinespace
Structural change and complexity & Change strength & Largest standardised mean shift across candidate splits. \\
Structural change and complexity & Linear fit error ratio & Linear one-step error relative to persistence. \\
Structural change and complexity & Turning behaviour ratio & Rate of direction changes in successive differences. \\
Structural change and complexity & Direction entropy & Entropy of adjacent change directions. \\
\addlinespace
Spectral structure & Spectral concentration & Detrended energy in the strongest frequency bins. \\
Spectral structure & Spectral entropy & Normalised spectral entropy. \\
Spectral structure & Stability & Variance of standardised block means. \\
\addlinespace
Missingness & Longest missing block & Longest consecutive missing segment. \\
Periodicity & Recent period strength & Strongest recent detrended periodic component. \\
Covariates & Strongest covariate relation & Largest absolute correlation between the target and any covariate. \\
Covariates & Covariate relation stability & Agreement between correlations of the target with covariates in early and late segments. \\
Intermittency & Recent event rate change & Change in recent versus historical nonzero rate. \\
\bottomrule
\end{tabular}
\end{table}

\clearpage
\section{Prompt Templates}
\label{app:prompt-templates}

All controller variants use the same shared system and user templates.
Memory Only and FASE additionally include the episodic memory instruction and retrieved memory block, whereas Policy Only and FASE include the online policy instruction and ranking block.
The definitions of the statistical features in Table~\ref{tab:statistical-features} are prepended to every user message.
The serialised values inserted into the templates vary across forecasting instances, but the surrounding instructions remain fixed.

\paragraph{Shared system instruction.}
\begingroup
\small
\begin{verbatim}
You are selecting a forecasting model for a time-series prediction task.
Choose one model from available_model_ids that you expect to achieve the
best point-forecast accuracy for the current case.
The supplied statistics describe the historical observations available at
the current forecast origin.
Use these statistics to understand the sequence's trend, periodicity,
variability, missingness, and other measured characteristics.
Consider how these characteristics affect forecasting over the requested
prediction horizon.
Use the model capability descriptions to assess each model's suitability
for the current input, including its supported variables, covariates,
missing-value handling, and context length.
Base your selection on the information provided for this case and your
knowledge of time-series forecasting.
Interpret null as an unavailable statistical measurement or a missing
historical observation.
\end{verbatim}
\endgroup

\paragraph{Episodic memory instruction.}
The following instruction is appended when episodic memory is enabled.
\begingroup
\small
\begin{verbatim}
The retrieved memories provide additional observations from
previous cases, including their historical statistics and the
performance of every evaluated model.
Consider these observations together with the current statistics, model
capabilities, and your time-series forecasting knowledge.
When assessing each available model for the current case, examine
how that model's performance varied across the retrieved cases as
their historical statistics varied.
Use these past relationships to refine how well you expect each model to
perform for the current case.
Lower mase and lower mase_rank indicate better point-forecast
performance.
Lower quantile_loss_rank and lower
mean_weighted_sum_quantile_loss indicate better probabilistic
performance.
Use the probabilistic results as supporting information when the
point-forecast evidence for multiple models is similar.
\end{verbatim}
\endgroup

\paragraph{Online policy instruction.}
Policy Only uses the following instruction.
\begingroup
\small
\begin{verbatim}
Use the online ranking as additional information when estimating the
performance of each available model for the current case. Consider it
together with the current statistics and the model capabilities. Compare
all six models and select the model with the best expected point-forecast
accuracy.
\end{verbatim}
\endgroup

When episodic memory and the online policy are both enabled, FASE uses the following version.
\begingroup
\small
\begin{verbatim}
Use the online ranking as additional information when estimating the
performance of each available model for the current case. Consider it
together with the current statistics, the retrieved cases, the model
capabilities, and the probabilistic forecasting results. Compare
all six models and select the model with the best expected
point-forecast accuracy.
\end{verbatim}
\endgroup

Every system message ends with the following output instruction.
\begingroup
\small
\begin{verbatim}
Return your selection as a JSON object with the field
"selected_model_id".
\end{verbatim}
\endgroup

\paragraph{User template.}
\begingroup
\small
\begin{verbatim}
[STATISTICAL FEATURE DEFINITIONS]

The following statistics describe the historical observations for the
current case.
<statistics>
[CURRENT STATISTICS]
</statistics>

[RETRIEVED MEMORY BLOCK, IF ENABLED]
[ONLINE MODEL RANKING BLOCK, IF ENABLED]

The following forecasting models are available for this prediction.
<available_model_ids>
[AVAILABLE MODEL IDENTIFIERS]
</available_model_ids>

The following model cards describe the supported inputs and
outputs of the forecasting models.
<model_capabilities>
[MODEL CAPABILITY DESCRIPTIONS]
</model_capabilities>
\end{verbatim}
\endgroup

\paragraph{Statistical feature definitions.}
The following guide replaces the first placeholder in the user template.
Line breaks are added only for presentation.
\begingroup
\small
\begin{verbatim}
The following definitions apply to both the current statistics and the
retrieved cases' statistics.
L is task.statistics_history_length, the number of historical time
positions used for statistics, up to the last 15360 positions including
the latest history.
task.forecast_horizon is the number of future positions to predict, and
task.frequency describes the sampling frequency.
task.past_covariate_count describes the number of historical covariate
channels.
Recent trend, level, and scale statistics use the last max(4, L // 4)
positions, limited by the available history.
Recent periodicity uses the last max(12, L // 3) positions, and recent
nonzero activity uses the last max(8, L // 4) positions, each limited by
the available history.
The separately supplied recent_history observations cover the last
forecast_horizon historical positions when that input is enabled.
quality_profile.missing_ratio is the fraction of missing positions across
all historical target values.
quality_profile.zero_ratio is the fraction of observed historical target
values equal to zero.
global_recent_dynamics.global_trend_strength is the fitted historical
slope multiplied by L and divided by the historical standard deviation,
with positive values indicating an upward trend and negative values
indicating a downward trend, summarized by the median across valid
channels.
global_recent_dynamics.recent_trend_strength is the fitted recent slope
multiplied by the recent segment length and divided by the historical
standard deviation, summarized by the median across valid channels.
global_recent_dynamics.recent_level_shift_strength is the absolute
difference between recent and preceding historical means divided by the
historical standard deviation, summarized by the maximum across valid
channels.
global_recent_dynamics.recent_scale_change_strength is the absolute
difference between recent and preceding historical standard deviations
divided by their sum, summarized by the maximum across valid channels.
change_regime_analysis.change_strength is the largest absolute difference
between means before and after a split in approximately the middle 20% to
80% of history, divided by the historical standard deviation and
maximized across valid channels.
nonlinear_complexity.linear_fit_error_ratio compares the mean squared
residual of a fitted one-step linear model with that of repeating the
previous observation, using adjacent observed pairs and averaging across
valid channels.
nonlinear_complexity.turning_behavior_ratio is the fraction of observed
consecutive triples whose two successive changes have different signs,
including zero as a sign and averaging across valid channels.
nonlinear_complexity.direction_entropy is the binary entropy of upward
versus non-upward changes in adjacent observed pairs, averaged across
valid channels.
dependence_screen.spectral_concentration is the fraction of
nonzero-frequency energy in the three strongest frequency bins after
removing a quadratic trend from complete historical channels, averaged
across valid channels.
background.entropy is normalized power-spectrum entropy of standardized
complete nonconstant historical channels, with higher values indicating
more dispersed spectral energy, summarized by the median across valid
channels.
background.stability is the sample variance of block means in
standardized complete nonconstant historical channels, with higher values
indicating greater changes in historical mean levels, summarized by the
median across valid channels.
The block width for background.stability is task.seasonal_period when
greater than 1, otherwise 10, and at least two complete blocks are
required.
missingness_structure.longest_missing_block is the longest consecutive
historical missing block across target channels, measured in time steps
and provided when missing values occur.
time_varying_periodicity.recent_period_strength is the strongest
nonzero-frequency energy fraction in the recent periodicity segment after
quadratic detrending, averaged across valid complete channels and
provided when L is at least 36 and a candidate period is detected.
covariate_relationship.strongest_covariate_relation is the maximum
absolute Pearson correlation between historical targets and covariates
over time-aligned observed pairs, provided when covariates are present.
covariate_relationship.covariate_relation_stability compares
target-covariate correlations in the first and second halves of history
using max(0, 1 - abs(first-half correlation - second-half correlation)),
averaged across valid target-covariate pairs.
intermittent_structure.recent_event_rate_change is the absolute
difference between recent and whole-history observed nonzero fractions,
maximized across valid channels and provided when L is at least 8 and the
historical zero ratio is at least 0.45.
null denotes an unavailable statistical value, and failed_tool_ids
identifies statistical tools whose computation failed.
\end{verbatim}
\endgroup

\paragraph{Model capability descriptions.}
The following object replaces the model capability placeholder in every
controller variant.
\begingroup
\small
\begin{verbatim}
{
  "seasonal_naive": {
    "multivariate_input": "independent_channels",
    "past_covariates": "ignored",
    "future_covariates": "ignored",
    "missing_values": "requires_linear_interpolation",
    "context_limit": {
      "type": "no_adapter_limit",
      "maximum_history_points": null
    }
  },
  "chronos2": {
    "multivariate_input": "joint_channels",
    "past_covariates": "ignored",
    "future_covariates": "ignored",
    "missing_values": "native",
    "context_limit": {
      "type": "finite",
      "maximum_history_points": 8192
    }
  },
  "timesfm_2p5": {
    "multivariate_input": "independent_channels",
    "past_covariates": "ignored",
    "future_covariates": "ignored",
    "missing_values": "requires_linear_interpolation",
    "context_limit": {
      "type": "finite",
      "maximum_history_points": 15360
    }
  },
  "moirai_2_small": {
    "multivariate_input": "independent_channels",
    "past_covariates": "ignored",
    "future_covariates": "ignored",
    "missing_values": "native",
    "context_limit": {
      "type": "finite",
      "maximum_history_points": 4000
    }
  },
  "tirex2_gifteval_zs": {
    "multivariate_input": "joint_channels",
    "past_covariates": "consumed",
    "future_covariates": "consumed",
    "missing_values": "native",
    "context_limit": {
      "type": "finite",
      "maximum_history_points": 8192
    }
  },
  "toto_2_313m": {
    "multivariate_input": "joint_channels",
    "past_covariates": "consumed",
    "future_covariates": "ignored",
    "missing_values": "native",
    "context_limit": {
      "type": "finite",
      "maximum_history_points": 4096
    }
  }
}
\end{verbatim}
\endgroup

The optional episodic memory and online policy blocks are formatted as follows.
\begingroup
\small
\begin{verbatim}
The following retrieved cases contain historical statistics and observed
model performance.
<memories>
[RETRIEVED CASES AND MODEL OUTCOMES]
</memories>

<online_model_ranking>
The following ranking is generated by an online model ranker. It is
updated over time using the observed point-forecast performance of
the six models on previously completed cases from this task.

It provides an additional estimate of the relative point-forecast
performance of the six models for the current case:

[ORDERED MODEL IDENTIFIERS]
</online_model_ranking>
\end{verbatim}
\endgroup

\clearpage
\section{Per-Configuration Results}
\label{app:per-configuration-results}

Tables~\ref{tab:per-config-mase} and~\ref{tab:per-config-crps} report normalised MASE and CRPS for each evaluated configuration.
All values are normalised by the corresponding score of Seasonal Naive, which is therefore omitted from the tables.

\begin{table}[H]
\centering
\scriptsize
\setlength{\tabcolsep}{2.2pt}
\caption{Normalised MASE for each evaluated GIFT-Eval configuration. Lower values indicate better point forecasting performance.}
\label{tab:per-config-mase}
\begin{tabular*}{\textwidth}{@{\extracolsep{\fill}}llccccccc@{}}
\toprule
Dataset & Setting & Chronos-2 & \shortstack{TimesFM\\2.5} & \shortstack{Moirai-2.0-\\R-Small} & TiRex-2 & \shortstack{Toto-2.0-\\313m} & \shortstack{Uniform\\Ensemble} & \textbf{FASE} \\
\midrule
bitbrains\_fast\_storage & 5T / medium & 0.777 & 0.815 & 0.813 & 0.797 & 0.751 & 0.777 & 0.732 \\
 & 5T / long & 0.762 & 0.795 & 0.807 & 0.790 & 0.719 & 0.758 & 0.709 \\
 & H / short & 0.754 & 0.818 & 0.822 & 0.823 & 0.717 & 0.774 & 0.723 \\
\addlinespace
bitbrains\_rnd & 5T / medium & 0.959 & 0.969 & 0.969 & 0.967 & 0.949 & 0.958 & 0.948 \\
 & 5T / long & 0.933 & 0.948 & 0.952 & 0.945 & 0.926 & 0.934 & 0.921 \\
 & H / short & 0.964 & 0.969 & 0.962 & 0.979 & 0.956 & 0.962 & 0.950 \\
\addlinespace
bizitobs\_service & 10S / short & 0.597 & 0.593 & 0.713 & 0.625 & 0.581 & 0.591 & 0.572 \\
car\_parts\_with\_missing & M / short & 0.694 & 0.698 & 0.688 & 0.695 & 0.729 & 0.696 & 0.699 \\
\addlinespace
electricity & D / short & 0.692 & 0.704 & 0.715 & 0.707 & 0.718 & 0.683 & 0.685 \\
 & H / medium & 0.761 & 0.760 & 0.779 & 0.763 & 0.766 & 0.744 & 0.744 \\
 & H / long & 0.781 & 0.752 & 0.828 & 0.769 & 0.798 & 0.763 & 0.756 \\
 & W-FRI / short & 0.680 & 0.716 & 0.794 & 0.727 & 0.775 & 0.713 & 0.695 \\
\addlinespace
hierarchical\_sales & D / short & 0.654 & 0.657 & 0.659 & 0.656 & 0.660 & 0.652 & 0.662 \\
hospital & M / short & 0.839 & 0.824 & 0.831 & 0.830 & 0.817 & 0.816 & 0.825 \\
\addlinespace
kdd\_cup\_2018\_with\_missing & D / short & 0.806 & 0.796 & 0.829 & 0.792 & 0.796 & 0.797 & 0.787 \\
 & H / medium & 0.780 & 0.726 & 0.788 & 0.739 & 0.735 & 0.724 & 0.717 \\
 & H / long & 0.804 & 0.755 & 0.802 & 0.759 & 0.767 & 0.750 & 0.746 \\
\addlinespace
LOOP\_SEATTLE & 5T / long & 0.785 & 0.676 & 0.742 & 0.653 & 0.661 & 0.665 & 0.607 \\
 & D / short & 0.516 & 0.498 & 0.516 & 0.500 & 0.510 & 0.498 & 0.489 \\
 & H / medium & 0.607 & 0.540 & 0.635 & 0.499 & 0.616 & 0.563 & 0.499 \\
 & H / long & 0.560 & 0.532 & 0.585 & 0.497 & 0.567 & 0.534 & 0.496 \\
\addlinespace
m4\_daily & D / short & 1.031 & 1.006 & 0.937 & 1.003 & 0.970 & 0.969 & 0.941 \\
M\_DENSE & H / short & 0.526 & 0.517 & 0.533 & 0.527 & 0.509 & 0.510 & 0.481 \\
restaurant & D / short & 0.678 & 0.678 & 0.692 & 0.673 & 0.689 & 0.667 & 0.676 \\
\addlinespace
solar & 10T / short & 0.720 & 0.982 & 0.595 & 0.931 & 0.776 & 0.741 & 0.466 \\
 & 10T / medium & 0.855 & 0.908 & 1.078 & 0.889 & 0.956 & 0.870 & 0.709 \\
 & 10T / long & 0.916 & 0.979 & 1.161 & 0.928 & 0.899 & 0.918 & 0.767 \\
 & H / short & 1.035 & 0.943 & 0.924 & 0.435 & 0.924 & 0.788 & 0.425 \\
\addlinespace
SZ\_TAXI & 15T / short & 0.711 & 0.712 & 0.715 & 0.712 & 0.709 & 0.708 & 0.711 \\
\bottomrule
\end{tabular*}
\end{table}

\clearpage
\begin{table}[H]
\centering
\scriptsize
\setlength{\tabcolsep}{2.2pt}
\caption{Normalised CRPS for each evaluated GIFT-Eval configuration. Lower values indicate better probabilistic forecasting performance.}
\label{tab:per-config-crps}
\begin{tabular*}{\textwidth}{@{\extracolsep{\fill}}llccccccc@{}}
\toprule
Dataset & Setting & Chronos-2 & \shortstack{TimesFM\\2.5} & \shortstack{Moirai-2.0-\\R-Small} & TiRex-2 & \shortstack{Toto-2.0-\\313m} & \shortstack{Uniform\\Ensemble} & \textbf{FASE} \\
\midrule
bitbrains\_fast\_storage & 5T / medium & 0.519 & 0.634 & 0.579 & 0.497 & 0.479 & 0.514 & 0.607 \\
 & 5T / long & 0.636 & 0.706 & 0.685 & 0.584 & 0.500 & 0.593 & 0.666 \\
 & H / short & 0.798 & 0.756 & 0.601 & 0.654 & 0.572 & 0.651 & 0.679 \\
\addlinespace
bitbrains\_rnd & 5T / medium & 1.015 & 0.722 & 0.508 & 0.533 & 0.472 & 0.640 & 0.533 \\
 & 5T / long & 0.861 & 0.632 & 0.484 & 0.505 & 0.454 & 0.573 & 0.881 \\
 & H / short & 0.830 & 0.490 & 0.538 & 0.480 & 0.472 & 0.546 & 0.548 \\
\addlinespace
bizitobs\_service & 10S / short & 0.264 & 0.255 & 0.339 & 0.278 & 0.257 & 0.261 & 0.244 \\
car\_parts\_with\_missing & M / short & 0.558 & 0.547 & 0.544 & 0.553 & 0.634 & 0.559 & 0.595 \\
\addlinespace
electricity & D / short & 0.528 & 0.519 & 0.520 & 0.524 & 0.544 & 0.505 & 0.492 \\
 & H / medium & 0.586 & 0.577 & 0.631 & 0.586 & 0.605 & 0.568 & 0.571 \\
 & H / long & 0.570 & 0.531 & 0.635 & 0.548 & 0.594 & 0.545 & 0.542 \\
 & W-FRI / short & 0.478 & 0.525 & 0.669 & 0.507 & 0.565 & 0.520 & 0.510 \\
\addlinespace
hierarchical\_sales & D / short & 0.331 & 0.331 & 0.333 & 0.333 & 0.335 & 0.329 & 0.348 \\
hospital & M / short & 0.863 & 0.811 & 0.829 & 0.827 & 0.791 & 0.801 & 0.832 \\
\addlinespace
kdd\_cup\_2018\_with\_missing & D / short & 0.565 & 0.562 & 0.583 & 0.558 & 0.566 & 0.563 & 0.581 \\
 & H / medium & 0.597 & 0.563 & 0.666 & 0.571 & 0.576 & 0.567 & 0.562 \\
 & H / long & 0.510 & 0.480 & 0.566 & 0.476 & 0.493 & 0.484 & 0.479 \\
\addlinespace
LOOP\_SEATTLE & 5T / long & 0.672 & 0.597 & 0.685 & 0.569 & 0.571 & 0.575 & 0.531 \\
 & D / short & 0.417 & 0.396 & 0.418 & 0.405 & 0.414 & 0.401 & 0.392 \\
 & H / medium & 0.389 & 0.342 & 0.427 & 0.306 & 0.388 & 0.353 & 0.308 \\
 & H / long & 0.318 & 0.305 & 0.351 & 0.275 & 0.319 & 0.300 & 0.276 \\
\addlinespace
m4\_daily & D / short & 0.927 & 0.891 & 0.823 & 0.881 & 0.852 & 0.853 & 0.830 \\
M\_DENSE & H / short & 0.466 & 0.465 & 0.480 & 0.476 & 0.460 & 0.456 & 0.435 \\
restaurant & D / short & 0.387 & 0.388 & 0.394 & 0.385 & 0.394 & 0.381 & 0.387 \\
\addlinespace
solar & 10T / short & 0.467 & 0.654 & 0.408 & 0.615 & 0.494 & 0.478 & 0.454 \\
 & 10T / medium & 0.445 & 0.520 & 0.630 & 0.495 & 0.510 & 0.464 & 0.442 \\
 & 10T / long & 0.431 & 0.491 & 0.606 & 0.455 & 0.445 & 0.439 & 0.429 \\
 & H / short & 0.587 & 0.569 & 0.579 & 0.261 & 0.536 & 0.461 & 0.307 \\
\addlinespace
SZ\_TAXI & 15T / short & 0.649 & 0.650 & 0.652 & 0.650 & 0.647 & 0.645 & 0.648 \\
\bottomrule
\end{tabular*}
\end{table}

\end{document}